\documentclass[11pt]{article}

\usepackage[final]{acl}

\usepackage{times}
\usepackage{latexsym}

\usepackage[T1]{fontenc}

\usepackage[utf8]{inputenc}

\usepackage{microtype}

\usepackage{inconsolata}

\usepackage{graphicx}
\usepackage{xspace,mfirstuc,tabulary}
\usepackage{xcolor}         
\usepackage{array}
\usepackage{multirow}
\usepackage{makecell}
\usepackage{amsmath}
\usepackage{adjustbox}
\usepackage{tabularx}
\usepackage{wrapfig} 
\usepackage{enumitem}
\usepackage{booktabs}
\usepackage{subcaption}
\usepackage{amssymb}

\title{MATCHA \raisebox{-0.2em}{\includegraphics[height=1.1em]{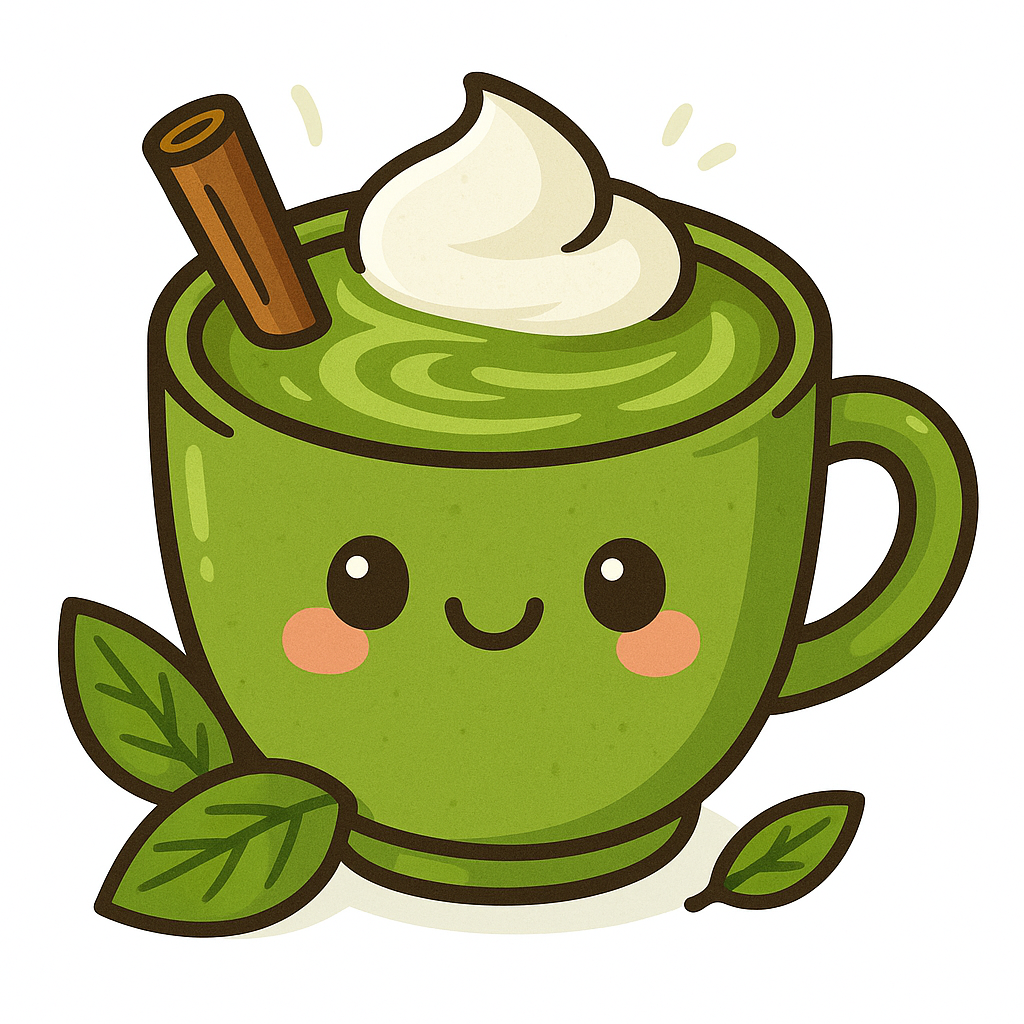}}: Matching Text via Contrastive Semantic Alignment}

\author{
    Siran Li \hspace{2em} Ece Sena Etoglu \hspace{2em} Carsten Eickhoff \hspace{2em} Seyed Ali Bahrainian \\
    University of Tübingen \\
    \texttt{siran.li@uni-tuebingen.de, ece.etoglu@guest.uni-tuebingen.de} \\
    \texttt{\{carsten.eickhoff, seyed-ali.bahreinian\}@uni-tuebingen.de} 
}

\begin{document}
\maketitle
\begin{abstract}
Reliable evaluation is essential for understanding large language model (LLM) performance, yet today's go-to metrics, namely token-overlap scores (e.g., ROUGE) and embedding-based measures (e.g., BERTScore), often misjudge semantic similarity of documents. Our study shows that both token-overlap metrics and embedding-based metrics routinely assign nearly identical scores to texts that directly contradict each other, thereby potentially masking fundamental errors. We introduce MATCHA, an automatic metric that jointly rewards semantic agreement with a reference and penalizes contradictions. MATCHA employs a dual-view perspective that measures (i) proximity to the gold text and (ii) distance from an adversarially generated counterfactual contradiction. In eight public benchmarks, MATCHA outperforms popular metrics, compared with human annotations on question-answering, image caption generation, natural language inference, summarization, and semantic textual similarity tasks. On the TruthfulQA dataset (i.e., a dataset without a training set, where no embedding-based metrics could locally train on), this improvement in terms of matching texts with a reference reaches 18.38\% over ROUGE-L and 20.82\% over BERTScore. Both quantitative comparison and qualitative human assessments confirm the efficacy and validity of MATCHA and uncover fundamental weaknesses in pre-existing metrics. Compared with 23 embedding models, including top state-of-the-art ones, used as a metric similar to BERTScore, MATCHA remains the most accurate in distinguishing correct from incorrect statements solely based on a reference. Our code and metric are publicly available~\footnote{\url{https://github.com/Siran-Li/MATCHA}}. 
\end{abstract}

\section{Introduction}
\label{sec:introduction}

In text generation research and applications, evaluation constitutes the main bottleneck in assessing genuine progress, mainly because consistently and objectively measuring the quality, relevance, or correctness of generated outputs remains a highly challenging task, particularly in knowledge-intensive and multimodal settings~\cite{mehri-eskenazi-2020-usr, pang2020towards, rei-etal-2020-comet, 11201031}. Various applications of text generation attract thousands of publications each year at prestigious conferences such as ACL, EMNLP, NeurIPS, and others. Despite rigorous peer review processes, reliably evaluating the validity of the evaluation sections in these research papers remains a challenge for both reviewers and post-acceptance readers~\cite{sai2022survey}. Although substantial effort is directed toward ensuring the robustness of evaluation methodologies, an essential, yet overlooked, concern is whether popular evaluation metrics themselves are adequate~\cite{reiter2018structured}, or whether shifts in the field have rendered historically significant metrics increasingly inadequate~\cite{dai-etal-2024-critical}.

\begin{table*}[t!]
\centering
\begin{minipage}[t]{0.48\textwidth}
\centering
\setlength\tabcolsep{4pt}
\scalebox{0.65}{\renewcommand{\arraystretch}{1.3}
\begin{tabular}{p{4.5cm}|l|cc}
\toprule
\multirow{2}{*}{\textbf{Example}} & \multirow{2}{*}{\textbf{Metric}} & \multirow{2}{*}{\textbf{Correct}} & \multirow{2}{*}{\textbf{Incorrect}} \\
& & & \\\midrule
\multirow{3}{*}{\parbox{4.5cm}{\textbf{Reference}: Growth is not affected by caffeine consumption.}} 
    & RL-F1     & 28.57 & 15.38 \\
    & METEOR    & 36.51 & 7.25 \\
    & EmbSim    & 81.00    & 77.19 \\
\multirow{2}{*}{\parbox{4.5cm}{\textbf{Correct}: Drinking coffee does not affect your growth.}} 
    & BERTScore & 82.00    & 77.20 \\
    & BLEURT    & -34.88 & -44.74 \\
\multirow{2}{*}{\parbox{4.5cm}{\textbf{Incorrect}: Drinking coffee will stunt your growth.}} 
    & SimCSE    & 88.88 & 75.21 \\
    & \textbf{Human Rating} & \textcolor{green}{\textbf{A (best)}} & \textcolor{red}{\textbf{E (worst)}} \\
\bottomrule
\end{tabular}}

\vspace{1ex}
\caption{Comparison of evaluation metrics and human judgment for correct vs. incorrect pairs. RL-F1 denotes ROUGE-L F1. Human Rating refers to judgments from three annotators on a five-point A–E scale.}\label{tab:intro-example}
\end{minipage}%
\hfill
\begin{minipage}[t]{0.48\textwidth}
\centering
\setlength\tabcolsep{5pt}
\scalebox{0.65}{\renewcommand{\arraystretch}{1.3}
\begin{tabular}{l|cc|cc}
\toprule
    \multirow{2}{*}{\centering\makecell[c]{\textbf{Metric}}} &  \multicolumn{2}{c|}{\textbf{MultiNLI}} & \multicolumn{2}{c}{\textbf{TruthfulQA}} \\ 
    ~ & \textbf{(Corr, Incorr)} & \textbf{$N\Delta$} & \textbf{(Corr, Incorr)} & \textbf{$N\Delta$} \\ \midrule
    \textbf{RL-F1} & (38.07, 30.38) & 7.69 & (40.72, 35.78) & 4.94 \\ 
    \textbf{METEOR} & (35.03, 26.03) & 9.00 & (42.08, 33.60) & 8.48 \\ 
    \textbf{EmbSim} & (71.59, 56.79) & 7.40 & (70.77, 63.11) & 3.83 \\ 
    \textbf{BERTScore} & (84.06, 80.62) & 3.44 & (83.80, 81.29) & 2.51 \\ 
    \textbf{BLEURT} & (-39.16, -75.55) & 18.20 & (-29.73, -63.46) & 16.87 \\ 
    \textbf{SimCSE} & (76.29, 49.55)  & 13.37 & (68.98, 55.77)  & 6.61 \\ 
    \textbf{MAUVE} & (56.01, 49.20) & 6.81 & (96.81, 86.60) & 10.21 \\ \bottomrule
\end{tabular}}
\vspace{1ex}
\caption{Average similarity scores for correct and incorrect pairs on two datasets. $N\Delta$ denotes the correct–incorrect gap after rescaling to $[0, 1]$.}\label{tab:intro-comparison}
\end{minipage}
\end{table*}

Evaluation metrics for text generation applications typically fall into two main categories: (1) Lexical overlap matching methods, which evaluate the quality of generated texts by counting overlapping textual units, such as n-grams, word sequences, or word pairs, between the generated candidate outputs and human-annotated references. Notable examples of this category include widely-used metrics such as ROUGE-N, ROUGE-L~\cite{lin-2004-rouge}, and Meteor~\cite{banerjee2005meteor}. (2) Model-based evaluation methods, which measure semantic similarity by mapping texts onto continuous embedding spaces. Prominent metrics in this category include BERTScore~\cite{Zhang2020BERTScore}, BLEURT~\cite{sellam2020bleurt}, SimCSE~\cite{gao-etal-2021-simcse}, and MAUVE~\cite{pillutla-etal:mauve:neurips2021}. Although metrics such as ROUGE, originating in the early 2000s, initially served evaluations in extractive summarization (i.e., extracting exact sentences from a document that together form a summary) when text generation capabilities were limited within the NLP literature, they have persisted and adapted to contemporary systems~\cite{karpinska-etal-2022-demetr}. Conversely, newer metrics such as BERTScore~\cite{Zhang2020BERTScore} have emerged alongside the rise of Transformer-based models, rapidly gaining traction and becoming virtual standards for assessing text generation across various NLP tasks. However, in recent years, new datasets with contrastive samples have opened new pathways both for building new robust evaluation metrics, as well as assessing existing text generation evaluation metrics that compare a model-generated response against a human-written one~\cite{guan-etal-2021-openmeva}. For example, we will show that an evaluation metric such as BERTScore assigns very high similarity scores not only to two contradictory documents but also to two documents that are semantically unrelated\footnote{For the semantically unrelated experiment, please see results on COCO-Caption in Table~\ref{tab:avg_score}}. Table \ref{tab:intro-example} shows the scores assigned by various popular text generation metrics to two contradicting sentences, one being correct and the other incorrect. We observe that model-based metrics such as BERTScore, Embedding Cosine Similarity (EmbSim) computed using SentenceTransformer~\cite{reimers-2019-sentence}, and SimCSE assign very similar scores to both correct and incorrect sentences, and as a standalone metric, they are not sufficiently discriminative and are also hardly human-interpretable. This trend continues and worsens as we generalize to large test datasets such as MultiNLI~\cite{williams-etal-2018-broad} and TruthfulQA~\cite{lin-etal-2022-truthfulqa}, as demonstrated in Table~\ref{tab:intro-comparison}. It is noteworthy that a distribution-based metric such as MAUVE cannot reliably evaluate single examples, as the authors explained\footnote{https://krishnap25.github.io/mauve} and instead requires larger sample sizes such as entire test sets to evaluate, which further adds to its limitation of not being able to score-wise distinguish between correct and incorrect, as demonstrated in Table~\ref{tab:intro-example}.

Metrics such as BERTScore and MAUVE do not explicitly guard against Type I errors, i.e., false positives where two texts are judged similar even though they diverge significantly. Because these scores rely on dense embedding-space distances (where most well-formed sentences cluster fairly closely together)~\cite{gaorepresentation}, they routinely assign high similarity to almost any pair of coherent documents. This tendency masks important semantic mismatches, so relying on these metrics alone can yield an overly optimistic view of a system's faithfulness or factual accuracy.

In this paper, we study existing text generation evaluation metrics in terms of reliability and robustness and demonstrate their inherent limitations. Subsequently, we introduce a new evaluation metric, MATCHA (Matching Text via Contrastive Semantic Alignment), which does not suffer from these shortcomings.

\textbf{Our key contributions are:} (1) We present an extensive analysis of existing text generation evaluation metrics and compare them robustly. (2) We present MATCHA, a new text semantic matching metric which is the closest to human judgments as compared with pre-existing metrics. MATCHA consistently achieves the best separation of contradictory statements across diverse NLP evaluation datasets; MATCHA learns a dual-perspective of matching similar documents while detecting dissimilar/contradictory documents, providing sharper semantic boundaries; MATCHA offers fine-grained interpretability by highlighting meaningful token-level semantic differences; and MATCHA shows stronger correlation with human similarity ratings, as well as, GPT-4 compared with previous metrics, confirming its reliability for detailed semantic evaluation. (3) We compare MATCHA with 23 top embedding models and show that it outperforms all of these models in distinguishing correct and incorrect statements using a reference.

\section{Related Work: Evaluation Metrics for Natural Language Generation}\label{sec:related_work}

Automatic evaluation metrics are essential for assessing natural language generation (NLG) quality. Traditional approaches range from n-gram-based methods like BLEU~\cite{papineni2002bleu} and ROUGE~\cite{lin-2004-rouge} to neural models such as BERTScore~\cite{Zhang2020BERTScore} and BLEURT~\cite{sellam2020bleurt}. While these metrics offer valuable approximations, they often fall short in capturing semantic nuances, contextual relevance, and factual correctness. This section reviews key evaluation paradigms, highlights their shortcomings, and outlines the motivation for new alternatives.

\textbf{Lexical overlap metrics provide a fast baseline but lack semantic awareness.} BLEU and ROUGE compare n-gram overlap between generated and reference texts. Although efficient and widely adopted, they poorly reflect human preferences in tasks involving paraphrasing~\cite{bahrainian-etal-2024-text} or abstraction~\cite{liu-etal-2023-g, phy-etal-2020-deconstruct}. METEOR~\cite{banerjee2005meteor} improves this by incorporating synonymy and stemming, but its reliance on static lexical resources limits scalability across domains~\cite{dziri-etal-2019-evaluating}.

\textbf{Semantic similarity metrics improve alignment with human judgments but struggle with nuance and reliability.} Embedding-based methods leverage semantic similarity through pretrained models. EmbSim~\cite{reimers-2019-sentence}, BERTScore~\cite{Zhang2020BERTScore}, BLEURT~\cite{sellam2020bleurt}, and SimCSE~\cite{gao-etal-2021-simcse} achieve stronger alignment with human judgments. MAUVE~\cite{pillutla-etal:mauve:neurips2021} further assesses divergence between machine and human text distributions. However, these methods are often insensitive to factual errors, truncations, or adversarial inputs~\cite{he-etal-2023-blind, deutsch-etal-2022-limitations, goyal-durrett-2020-evaluating}. MAUVE also struggles with fine-grained evaluation, requiring large text samples~\cite{pimentel2023on}. Thus, we need task-aware, robustness-oriented evaluators that penalize factual errors and adversarial perturbations.

\textbf{Current evaluation metrics lack semantic precision, interpretability, and robustness.} Despite steady progress, current evaluation metrics for NLG often fall short in capturing critical aspects such as semantic contradiction, factual inaccuracy, and over-reliance on fluency, where outputs appear fluent and well-structured despite containing incorrect or unsupported information~\cite{howcroft-etal-2020-twenty, schmidtova-etal-2024-automatic-metrics, golovanevsky-etal-2025-pixels}. Studies have shown that many metrics reward surface-level similarity while failing to penalize outputs that are semantically incorrect or inconsistent with a reference~\cite{he-etal-2023-blind, gehrmann-etal-2021-gem, fabbri2021summeval}. This is particularly problematic in settings where detecting subtle false positives or contradictory statements is essential. Moreover, model-based and embedding-based methods frequently offer limited interpretability, making it difficult to diagnose errors or understand model decisions~\cite{jiang2024tigerscore}. As shown in Table~\ref{tab:intro-comparison}, these issues persist across domains and tasks, calling for metrics that are both semantically precise and diagnostically useful. To address these challenges, we introduce MATCHA, a semantic matching metric that combines contrastive learning with token-level alignment to deliver accurate, interpretable, and robust NLG evaluation.

\section{MATCHA}
\label{sec:methodology}

\begin{figure*}[t!]
    \centering
    \includegraphics[width=0.8\textwidth]{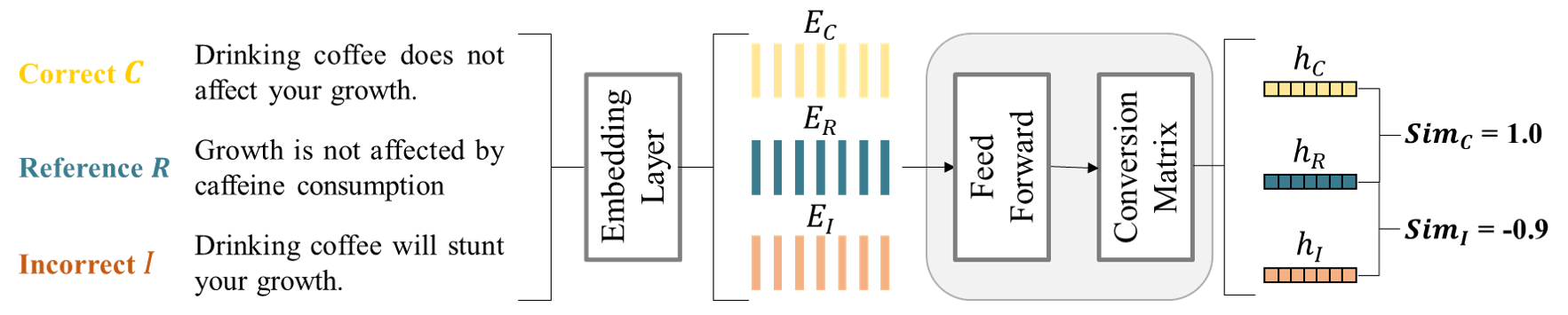}
\caption{
Overview of the MATCHA architecture. Each input Correct Candidate (C), Reference (R), and Incorrect Candidate (I) is first encoded into embeddings $\mathbf{E}$ via an embedding layer initialized from the GPT-2 decoder. A lightweight feed-forward projector with a conversion matrix $\mathbf{W}$ maps $\mathbf{E}$ into a shared semantic space, yielding context representations $\mathbf{h}$, which are compared via cosine similarity.
}\label{fig:diagram}
\end{figure*}

MATCHA is a contrastive matching metric that compares two documents by encoding them into contextualized representations, projecting them onto a contrastive-learned matching space, and computing their similarity. Inspired by the contrastive learning paradigm~\cite{chen2020simple, liu-etal-2022-brio, mi2022knowledge}, MATCHA encourages semantically similar inputs to have closer representations, while pushing dissimilar inputs apart. The overall architecture of MATCHA is illustrated in Figure~\ref{fig:diagram}. Our novel method consists of three main processing stages: Token Embedding, Representation Alignment, and Similarity Computation, offering both document-level matching and fine-grained token attribution.

\subsection{Token Embedding}\label{sub:token-embedding}

Given an input document of tokens $\mathbf{x} = (x_1, x_2, \ldots, x_L)$, we initialize the embedding layer using the pre-trained GPT-2 word embedding layer~\cite{radford2019language}. Each token $x_i$ is mapped to a $D$-dimensional vector:

\begin{equation}
\mathbf{E} = \mathrm{EmbeddingLayer}(\mathbf{x}) \quad \in \quad \mathbb{R}^{L \times D}
\end{equation}
where $L$ is the sequence length and $D$ is the hidden dimensionality. Only the embedding layer is transferred from the GPT-2 decoder; all subsequent layers are trained from scratch.

\subsection{Representation Projection}\label{sub:transformation}

To adapt the embeddings to our task-specific representation space, we project each token embedding into $N_c$ context vectors via a lightweight feedforward projection:
\begin{equation}
\mathbf{S} = \mathrm{FeedForward}(\mathbf{E}) \quad \in \quad \mathbb{R}^{N_c \times L \times D}
\end{equation}

Next, the context vectors are projected into a common representation space using a learned linear conversion matrix, $\mathbf{W} \in \mathbb{R}^{D \times D}$, defined as:
\begin{equation}
\mathbf{S'} = \mathbf{S} \times \mathbf{W}
\end{equation}
where the multiplication is applied across the last dimension.

Finally, a document-level representation is obtained by mean pooling across both token and context dimensions:
\begin{equation}
\mathbf{h} = \frac{1}{N_c L} \sum_{i=1}^{N_c} \sum_{j=1}^{L} \mathbf{S'}_{i,j}
\end{equation}
This yields a fixed-size representation $\mathbf{h} \in \mathbb{R}^{1 \times D}$ for each input document.

\subsection{Similarity Computation, Training, and Inference}\label{sub:cosine-similarity}

We describe how MATCHA computes similarity, learns contrastive data, and scores candidates.

\textbf{Similarity Score.} 
Given two documents with embeddings $\mathbf{h}_1$ and $\mathbf{h}_2$, their semantic similarity is measured using cosine similarity:
\begin{equation}
\mathrm{sim}(\mathbf{h}_1, \mathbf{h}_2) = \frac{\mathbf{h}_1 \cdot \mathbf{h}_2}{\|\mathbf{h}_1\| \|\mathbf{h}_2\|}
\end{equation}
where $\cdot$ denotes the dot product and $\|\cdot\|$ is the Euclidean norm.

\textbf{Training Objective.}
MATCHA is trained to maximize the distance between correct and incorrect candidates in the embedding space. Given a reference embedding $\mathbf{h}_R$, a positive candidate $\mathbf{h}_C$, and a negative candidate $\mathbf{h}_I$, their cosine similarities are:
\begin{equation}
\mathrm{sim}_C = \mathrm{sim}(\mathbf{h}_R, \mathbf{h}_C), \quad \mathrm{sim}_I = \mathrm{sim}(\mathbf{h}_R, \mathbf{h}_I)
\end{equation}

A margin-based contrastive loss enforces $\mathrm{sim}_C$ to be higher than $\mathrm{sim}_I$:

\begin{equation}
\mathcal{L} = \mathbb{E}\left[\max(0, m + \mathrm{sim}_I - \mathrm{sim}_C)\right]
\end{equation}
where $m$ is a fixed margin hyperparameter (default $m=1$). This encourages the model to pull reference–positive pairs closer together while pushing reference–negative pairs farther apart.

\textbf{Inference.}
At inference time, all model parameters are fixed. For a reference $R$ and a candidate $T$, MATCHA encodes both into document embeddings and computes their similarity score:
\begin{equation}
\mathrm{Score}(\text{R}, \text{T}) = \mathrm{sim}(\mathbf{h}_{\text{R}}, \mathbf{h}_{\text{T}})
\end{equation}

\begin{table*}[t!]
\setlength\tabcolsep{5pt}
\centering
\scalebox{0.58}{{\renewcommand{\arraystretch}{1.3}
\begin{tabular}{l|ll|ll|ll|ll|ll|ll}
\toprule
\multirow{2}{*}{\centering\makecell[c]{\textbf{Metric}}} &  \multicolumn{2}{c|}{\textbf{SNLI}} & \multicolumn{2}{c|}{\textbf{MultiNLI}} & \multicolumn{2}{c|}{\textbf{TruthfulQA}} & \multicolumn{2}{c|}{\textbf{Climate-Fever}} & \multicolumn{2}{c|}{\textbf{COCO-Caption}} & \multicolumn{2}{c}{\textbf{NEWTS}} \\ 
~ & \textbf{(Corr, Incorr)} & \textbf{$N\Delta$} & \textbf{(Corr, Incorr)} & \textbf{$N\Delta$} & \textbf{(Corr, Incorr)} & \textbf{$N\Delta$} & \textbf{(Corr, Incorr)} & \textbf{$N\Delta$} & \textbf{(Corr, Incorr)} & \textbf{$N\Delta$} & \textbf{(Corr, Incorr)} & \textbf{$N\Delta$}  \\ \midrule
\textbf{R1-F1} & (41.34, 28.58) & 12.76 & (45.32, 34.67) & 10.65 & (45.12, 38.87) & 6.25 & (26.78, 23.05) & 3.73 & (27.10, 12.49) & 14.61 & (19.18, 8.70) & 10.48  \\ 
\textbf{R2-F1} & (19.07, 10.26) & 8.81 & (22.74, 14.96) & 7.78 & (24.40, 21.53) & 2.87 & (7.55, 5.67) & 1.88 & (12.13, 1.60) & 10.53 & (2.91, 0.39) & 2.52 \\ 
\textbf{RL-F1} & (38.24, 26.44) & 11.80 & (38.07, 30.38) & 7.69 & (40.72, 35.78) & 4.94 & (19.62, 18.12) & 1.50 & (23.27, 11.32) & 11.95 & (12.74, 6.80) & 5.94 \\ 
\textbf{METEOR} & (30.08, 20.78) & 9.30 & (35.03, 26.03) & 9.00 & (42.08, 33.60) & 8.48 & (19.01, 15.33) & 3.68 & (13.19, 5.51) & 7.68 & (10.41, 3.74) & 6.67 \\ 
\textbf{EmbSim} & (65.70, 33.36) & 16.17 & (71.59, 56.79) & 7.40 & (70.77, 63.11) & 3.83 & (60.56, 55.74) & 2.41 & (64.33, 9.25) & 27.54 & (30.89, 5.69) & 12.61  \\ 
\textbf{BERTScore} & (83.05, 79.34) & 3.71 & (84.06, 80.62) & 3.44 & (83.80, 81.29) & 2.51 & (78.91, 77.74) & 1.17 & (83.01, 73.30) & 9.71 & (73.44, 66.14) & 7.30  \\ 
\textbf{BLEURT} & (-64.06, -99.63) & 17.79 & (-39.16, -75.55) & 18.20 & (-29.73, -63.46) & 16.87 & (-41.65, -53.76) & 6.06 & (-76.28, -124.05) & 23.89 & (-108.56, -135.69) & 13.57 \\ 
\textbf{SimCSE} & (69.47, 33.43) &	18.03 &	(76.29, 49.55) &	13.37 &	(68.98, 55.77) &	6.61 &	(65.63, 51.82) &	6.91 &	(70.95, 10.20) &	30.38 &	\textbf{(39.45, 11.53)} & \textbf{13.96} \\
\textbf{MAUVE} & (41.50, 47.67) & -6.17 & (56.01, 49.20) & 6.81 & (96.81, 86.60) & 10.21 & (70.95, 51.99) & 18.96 & (6.87, 6.16) & 0.71 & (1.64, 0.75) & 0.89 \\ 
\textbf{MATCHA} & \textbf{(71.14, 1.24)} & \textbf{34.95} & \textbf{(64.77, -9.76)} & \textbf{37.27} & \textbf{(62.50, 15.85)} & \textbf{23.33} & \textbf{(52.92, 12.57)} & \textbf{20.18} & \textbf{(81.11, 9.59)} & \textbf{35.75} & \textbf{(44.64, 16.53)} & \textbf{14.06}  \\
\bottomrule
\end{tabular}}}
\caption{Average similarity scores for correct and incorrect pairs across six benchmarks.  $N\Delta$ denotes the normalized correct-incorrect gap, with EmbSim, BLEURT, SimCSE, and MATCHA rescaled to $[0,1]$ via $(\text{score}+1)/2$. Bold indicates statistical significance over the next highest score, based on a paired t-test with a confidence of 95\%.}
\label{tab:avg_score}
\end{table*}

\begin{table}[t]
\centering
\scalebox{0.55}{{\renewcommand{\arraystretch}{1.3}
\begin{tabular}{l|cccccc}
\toprule
\textbf{Metric} & \textbf{SNLI} & \textbf{MultiNLI} & \textbf{TruthfulQA} & \textbf{Climate} & \textbf{COCO} & \textbf{NEWTS} \\
\midrule
\textbf{R1-F1}      & 57.69 & 58.15 & 55.04 & 37.11 & 33.67 & 33.71 \\
\textbf{R2-F1}      & 42.35 & 43.15 & 44.29 & 34.07 & 33.33 & 33.33 \\
\textbf{RL-F1}      & 54.67 & 50.90 & 51.53 & 34.48 & 33.33 & 33.52 \\
\textbf{METEOR}     & 46.68 & 50.51 & 55.20 & 36.93 & 33.33 & 33.43 \\
\textbf{EmbSim}     & 37.96 & 33.33 & 33.95 & 33.33 & 52.79 & \textbf{60.89} \\
\textbf{BERTScore}  & 33.33 & 33.33 & 33.33 & 33.33 & 33.33 & 33.33 \\
\textbf{BLEURT}     & 44.72 & 54.71 & 56.03 & 40.23 & 33.67 & 33.33 \\
\textbf{SimCSE}     & 39.20 & 33.71 & 34.65 & 33.33 & 55.04 & 49.76 \\
\textbf{MATCHA}     & \textbf{72.59} & \textbf{74.51} & \textbf{61.08} & \textbf{64.08} & \textbf{68.41} & 57.99 \\
\bottomrule
\end{tabular}}}
\caption{Macro-F1 classification performance under a fixed midpoint threshold across six benchmarks.}
\label{tab:f1_only}
\end{table}

\begin{figure*}[t!]
    \centering
    \includegraphics[width=1\textwidth]{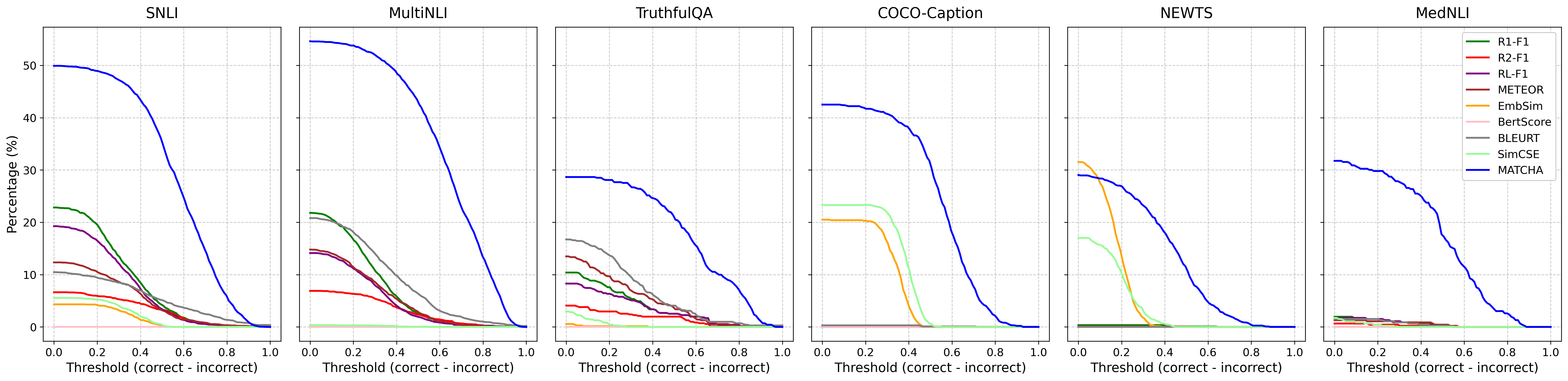}
    \caption{Comparison of different metrics' performance on six datasets. The y-axis shows the percentage of samples satisfying $(\text{Sim}_C - \text{Sim}_I) > \text{Threshold}$, with the middle point as the decision boundary. Metric scores are linearly transformed to $[0, 1]$.}\label{fig:metric_gap}
\end{figure*}

\section{Evaluation}
\label{sec:experiment}

In this section, we evaluate MATCHA across a broad range of tasks and demonstrate that it consistently outperforms existing metrics. Additional details on training, preprocessing, implementation details, and additional ablation analysis on data composition and optimization strategy are provided in Appendices~\ref{app:training-details},~\ref{app:eval-datasets},~\ref{app:implementation}, and~\ref{app:ablation-study}.

\subsection{Experimental Setup}\label{sub:setup}

\textbf{Baselines and Datasets.} We compare MATCHA against nine widely used metrics: ROUGE-1 (R1-F1), ROUGE-2 (R2-F1), ROUGE-L (RL-F1), METEOR, EmbSim, BERTScore, BLEURT, SimCSE, and MAUVE. To demonstrate the broad applicability of MATCHA, we evaluate it on eight datasets, including SNLI~\cite{bowman-etal-2015-large}, MultiNLI~\cite{williams-etal-2018-broad}, TruthfulQA~\cite{lin-etal-2022-truthfulqa}, Climate-Fever~\cite{diggelmann2020climatefever}, COCO-Caption~\cite{lin2015microsoft}, NEWTS~\cite{bahrainian-etal-2022-newts}, MedNLI~\cite{romanov-shivade-2018-lessons} and STS-B (English) dataset~\cite{cer-etal-2017-semeval}, spanning natural language inference (NLI), question-answering, factuality verification, image captioning, summarization, and semantic textual similarity (STS). 

\textbf{Evaluation Methods.} To assess the discriminative power of different metrics, we measure the average correct scores $\text{Sim}_C$ and incorrect scores $\text{Sim}_I$ and their gaps $N\Delta$, defined as the average of $(\text{Sim}_C - \text{Sim}_I)$ rescaled to $[0,1]$, the percentage of cases where $(\text{Sim}_C - \text{Sim}_I) > \text{Threshold}$, and the Wasserstein distance~\cite{villani2009wasserstein}. Further, we evaluate alignment with human judgment using Rank@1 (R@1), defined as the percentage of cases where a metric's score is closest to the human score, average Discounted Cumulative Gain (DCG)~\cite{jarvelin2002cumulated}, and concordance correlation coefficient (CCC)~\cite{lawrence1989concordance}. Unlike Pearson or Spearman, CCC captures not only rank correlation but also calibration, ensuring that metric scores align with human ratings in both ordering and scale. We also analyze token-level interpretability using Integrated Gradients~\cite{sundararajan2017axiomatic} to compute token-level attributions in Appendix~\ref{app:token-interpret}.

\subsection{Experimental Results}\label{sub:results}

To assess MATCHA's performance, we conduct four sets of experiments: (i) we compare it with existing metrics on contrastive datasets to distinguish texts with contrastive/irrelevant meanings (Section~\ref{subsub:comparison-results}); (ii) we evaluate MATCHA against 23 state-of-the-art embedding models, examining their semantic separation ability (Section~\ref{subsub:embedding_comparison}); (iii) we evaluate out-of-domain generalization (Section~\ref{subsub:unseen}); and 
(iv) via human evaluation (Section~\ref{subsub:human-eval}), we show that MATCHA aligns with human judgments more than the baselines.

\subsubsection{Metric Semantic Differentiation Power}\label{subsub:comparison-results}

We test the semantic differentiation power of MATCHA and baseline metrics on the following three aspects: (1) average similarity gap and macro-F1 classification performance (Table~\ref{tab:avg_score} and Table~\ref{tab:f1_only}), (2) proportion of samples with clear separation (Figure~\ref{fig:metric_gap}), and (3) Wasserstein distribution distance performance (Table~\ref{tab:wasser_scores}).

\textbf{Average similarity gap.} We measure the normalized similarity gap, which reflects how well a metric separates correct from incorrect samples generally (Table~\ref{tab:avg_score}). MATCHA achieves the highest gap on six datasets, substantially outperforming baseline metrics, with a gap of 34.95 on SNLI and 37.27 on MultiNLI, significantly ahead of the next best, SimCSE and BLEURT (18.03 and 18.20, respectively). On TruthfulQA, MATCHA outperforms BLEURT by +6.46\%. It further outperforms MAUVE by +1.22\% on Climate-Fever, +5.37\% on COCO-Caption, and +0.10\% on NEWTS. These results demonstrate MATCHA’s strong and consistent semantic differentiation ability, even on challenging out-of-domain tasks.

We additionally evaluate binary discrimination performance using macro-F1 under a fixed midpoint threshold (Table~\ref{tab:f1_only}). MATCHA delivers the strongest overall classification performance, achieving the highest macro-F1 on five of the six benchmarks. Compared with the best baseline on each dataset, MATCHA improves performance by +14.90 on SNLI, +16.36 on MultiNLI, +5.05 on TruthfulQA, +23.85 on Climate-Fever, and +13.37 on COCO-Caption. On NEWTS, MATCHA remains highly competitive at 57.99. These findings show that MATCHA’s larger semantic score margins translate into substantially more reliable decision boundaries for distinguishing correct from incorrect text pairs, while preserving robustness under challenging unseen-domain transfer settings.

\textbf{Margin between correct and incorrect scores.}
As shown in Figure~\ref{fig:metric_gap}, MATCHA consistently achieves larger margins across thresholds. In contrast, ROUGE, METEOR, and BERTScore drop off quickly, showing limited separation. EmbSim and SimCSE perform better than n-gram metrics but still lag behind MATCHA, especially on MultiNLI and TruthfulQA, where samples often share the same topic with high wording overlap, making contrastive distinctions more difficult than in other datasets.

\begin{table}[t!]
\setlength\tabcolsep{5pt}
\centering
\scalebox{0.60}{{\renewcommand{\arraystretch}{1.3}
\begin{tabular}{l|cccccc}
\toprule
\textbf{Metric} & \textbf{SNLI} & \textbf{MultiNLI} & \textbf{TruthfulQA} & \textbf{Climate} & \textbf{COCO} & \textbf{NEWTS} \\ \midrule
\textbf{R1-F1} & 12.77 & 10.65 & 6.26 & 3.87 & 14.61 & 10.48  \\ 
\textbf{R2-F1} & 8.81 & 7.77 & 2.98 & 1.88 & 10.53 & 2.52  \\ 
\textbf{RL-F1} & 11.80 & 7.69 & 4.95 & 1.66 & 11.95 & 5.94  \\ 
\textbf{METEOR} & 9.30 & 8.99 & 8.48 & 3.68 & 7.68 & 6.67  \\ 
\textbf{EmbSim} & 16.17 & 7.40 & 3.85 & 2.42 & 27.54 & 12.60  \\ 
\textbf{BERTScore} & 3.72 & 3.43 & 2.51 & 1.17 & 9.71 & 7.30  \\ 
\textbf{BLEURT} & 17.79 & 18.20 & 16.91 & 6.06 & 23.89 & 13.56  \\ 
\textbf{SimCSE} & 18.02 & 13.37 & 6.61 & 6.91 & 30.38 & 13.96  \\ 
\textbf{MATCHA} & \textbf{34.95} & \textbf{37.27} & \textbf{23.33} & \textbf{20.17} & \textbf{35.76} & \textbf{14.05}  \\ \bottomrule
\end{tabular}}}
\caption{Wasserstein distribution distance between correct and incorrect scores. Climate and COCO denote Climate-Fever and COCO-Caption, respectively.} 
\label{tab:wasser_scores}
\end{table}

\textbf{Wasserstein distribution distance performance.}
We evaluate separation between correct and incorrect scores using Wasserstein distribution distance, a non-parametric metric that captures how far apart two distributions are without assuming any specific shape. As shown in Table~\ref{tab:wasser_scores}, MATCHA consistently records the highest distances, with especially large margins on SNLI (34.95) and MultiNLI (37.27), nearly doubling the strongest baselines. It also surpasses the strongest baselines by +6.42\% on TruthfulQA, +13.26\% on Climate-Fever, and +5.38\% on COCO-Caption, while maintaining competitive performance on NEWTS, confirming its ability to avoid score clustering and provide more reliable separation between correct and incorrect samples.

Traditional metrics such as ROUGE, METEOR, and BERTScore consistently show low similarity gaps, weak separation margins, and poor macro-F1 discrimination performance. BLEURT assigns negative scores on average to both correct and incorrect cases, although it remains competitive on some datasets. EmbSim and SimCSE perform better than the previous metrics but still fall behind MATCHA, particularly in classification robustness and cross-domain generalization. MAUVE, which measures distributional similarity rather than direct semantic alignment, also fails to provide meaningful separation in SNLI and COCO-Caption. In contrast, MATCHA consistently delivers larger distinction gaps, stronger separation margins, higher Wasserstein distribution distances, and the best macro-F1 scores across datasets, establishing itself as a more reliable measure of semantic distinction.

\begin{figure*}[t!]
    \centering
    \includegraphics[width=1\textwidth]{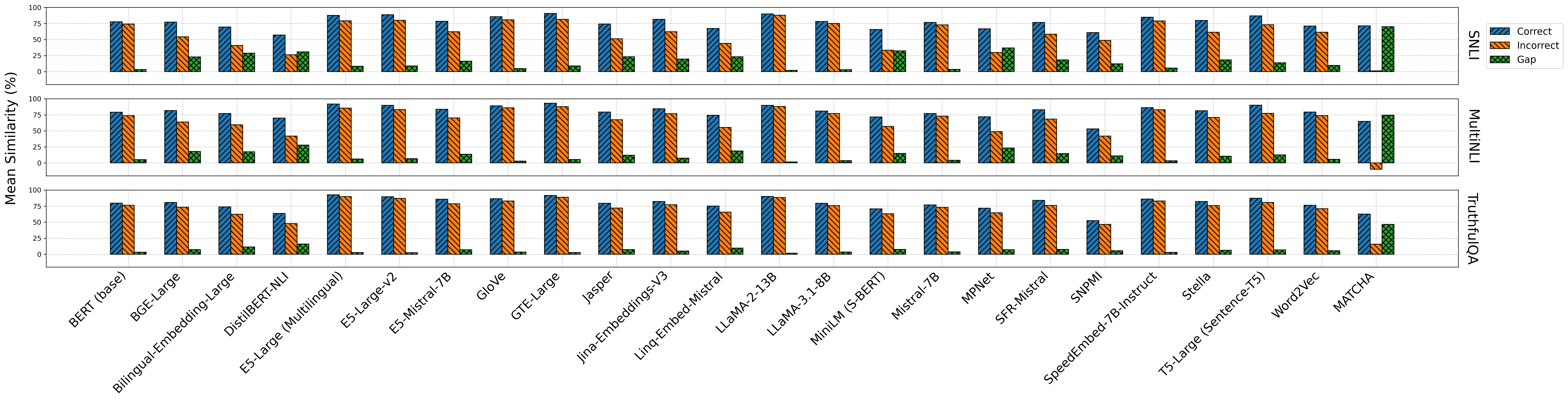}
    \caption{Average similarity scores and gaps between correct and incorrect pairs across embedding models on SNLI, MultiNLI, and TruthfulQA.}
\label{fig:bar_lmembed}
\end{figure*}

\subsubsection{Embedding-Based Semantic Separation}\label{subsub:embedding_comparison}

We evaluate MATCHA in the embedding space by comparing it with 23 embedding models, testing their ability to distinguish between correct and incorrect pairs using cosine similarity. The evaluated models form one of \textbf{the most comprehensive comparisons to date}, including top STS performers from the MTEB benchmark~\cite{muennighoff2023mteb}\footnote{\url{https://huggingface.co/spaces/mteb/leaderboard}, see the English leaderboard}, as well as baseline transformers and traditional word embeddings.

We include the following models: Bilingual-Embedding-Large~\cite{bilingual_embedding_large}, Jina-Embeddings-V3~\cite{gunther2024jina}, SFR-Embedding-Mistral~\cite{sfr_embedding_mistral}, Speed-Embedding-7B-Instruct~\cite{speed_embedding_7b_instruct}, Multilingual-E5-Large-Instruct~\cite{wang2024multilingual}, Linq-Embed-Mistral~\cite{linq_embed_mistral}, Jasper and Stella~\cite{novasearch2024jasper}, GTE-Large~\cite{gte_large}, and the instruction-tuned LLM E5-Mistral-7B-Instruct~\cite{e5_mistral_7b_instruct}.
Additionally, we evaluate widely used pre-trained sentence encoders such as MiniLM~\cite{wang2021minilmv2}, MPNet~\cite{song2020mpnet}, Sentence-T5~\cite{ni2022sentence}, E5-Large-v2~\cite{wang2023improving}, BGE-Large~\cite{bge_large_v1_5}, and BERT (base)~\cite{devlin2019bert}, along with DistilBERT-NLI~\cite{sanh2020distilbertdistilledversionbert} included as a relevant semantic baseline, due to its fine-tuning on NLI datasets. We also examine embeddings derived from LLMs such as LLaMA-2~\cite{touvron2023llama2}, LLaMA-3.1-8B~\cite{llama3.1-8b}, and Mistral-7B~\cite{jiang2023mistral}. 
Lastly, our study includes SNPMI~\cite{bahrainian2021self}, a co-occurrence-based embedding constructed from normalized pointwise mutual information (NPMI), alongside traditional word embedding baselines such as GloVe~\cite{pennington2014glove} and Word2Vec~\cite{mikolov2013efficient}. 

\textbf{Average similarity gap among 23 LM embeddings.}
Across SNLI, MultiNLI, and TruthfulQA, MATCHA produces the largest contrast between correct and incorrect scores (Figure~\ref{fig:bar_lmembed}). While most embeddings assign uniformly high similarities, yielding small gaps, MATCHA clearly separates the two and is the only model to give negative average similarity to incorrect samples (–9.76 on MultiNLI) while preserving high similarity for correct ones. This strong calibration enables MATCHA to overcome the common “over-similarity” problem in large pretrained embeddings~\cite{ethayarajh-2019-contextual}, making it a more human-aligned and trustworthy metric than baselines.

Based on Figures~\ref{fig:bar_lmembed} and the semantic separation curves in Appendix~\ref{app:lm-seperation}, MATCHA achieves uniquely large similarity gaps and robust margins, addressing the shortcomings of both general-purpose embeddings and fine-tuned STS/NLI models. Due to space constraints, we present results on NLI datasets and TruthfulQA, with other datasets showing similar patterns (Figure~\ref{fig:app_avg_lmgap}). These findings demonstrate that MATCHA establishes sharper semantic boundaries and delivers more reliable discriminative performance across diverse tasks.

\begin{table}[t]
\centering
\small
\scalebox{0.80}{{\renewcommand{\arraystretch}{1.3}
\begin{tabular}{l|cccc}
\toprule
\textbf{Metric} & \textbf{(Corr, Incorr)} & \textbf{N$\Delta$} & \textbf{macro-F1} & \textbf{Wasserstein} \\
\midrule
R1-F1      & (14.22, 11.65)   & 2.57  & 37.45 & 2.58 \\
R2-F1      & (4.66, 3.12)     & 1.54  & 34.14 & 1.54 \\
RL-F1      & (13.31, 11.08)   & 2.23  & 36.56 & 2.25 \\
METEOR     & (8.36, 6.62)     & 1.74  & 35.14 & 1.75 \\
EmbSim     & (46.14, 40.47)   & 2.84  & 33.43 & 2.85 \\
BERTScore  & (74.96, 74.17)   & 0.79  & 33.33 & 1.00 \\
BLEURT     & (-122.43, -132.13) & 4.85 & 34.94 & 4.85 \\
SimCSE     & (50.43, 37.00)   & 6.72  & 35.35 & 6.72 \\
MAUVE      & (1.29, 1.40)     & -0.11  & --    & -- \\
\textbf{MATCHA} & \textbf{(50.79, 14.24)} & \textbf{18.28} & \textbf{59.49} & \textbf{18.27} \\
\bottomrule
\end{tabular}}}
\caption{Performance on MedNLI, reporting average similarity scores for correct and incorrect pairs, $N\Delta$, macro-F1, and Wasserstein distance.}
\label{tab:mednli_generalization}
\end{table}

\subsubsection{Out-of-Domain Generalization}\label{subsub:unseen}

To evaluate MATCHA’s robustness under domain shift, we further test it on MedNLI~\cite{romanov-shivade-2018-lessons}, a clinical-domain natural language inference benchmark derived from real medical notes. MedNLI contains specialized medical terminology and domain-specific reasoning patterns that differ substantially from the general-domain datasets used during training, making it a strong out-of-distribution evaluation setting.

As shown in Table~\ref{tab:mednli_generalization}, MATCHA consistently outperforms all baselines across average score separation ($N\Delta$), macro-F1, and Wasserstein distance. In particular, MATCHA achieves the highest normalized separation gap (18.28), improving over the strongest baseline SimCSE by +11.56\%. The same trend holds for macro-F1 under a fixed midpoint threshold, where MATCHA reaches 59.49, outperforming the best baseline R1-F1 by +22.04\%. MATCHA also records the strongest Wasserstein distance (18.27), exceeding the best baseline SimCSE by +11.55\%. These findings demonstrate that MATCHA maintains robust semantic separation under substantial domain shift and generalizes effectively to domains with unseen vocabulary and specialized reasoning requirements.

While threshold-based discrimination can suffice for binary decision making, large and well-calibrated score margins are especially desirable for evaluation metrics because they improve robustness and interpretability. In practice, evaluation metrics are often used across datasets, models, and tasks without a fixed or carefully tuned threshold; small score gaps make them highly sensitive to noise, calibration shifts, and domain changes. As shown in our experiments, metrics with narrow margins frequently assign similarly high scores to both correct and contradictory candidates, masking fundamental semantic errors. In contrast, MATCHA’s larger margins provide more stable separation, stronger alignment with human judgments, and more reliable comparisons across models and settings.

\begin{table}[t!]
\setlength\tabcolsep{5pt}
\centering
\scalebox{0.6}{{\renewcommand{\arraystretch}{1.3}
\begin{tabular}{cccccccccc}
\toprule
R1-F1 & R2-F1 & RL-F1 & MTR & EmbSim & BSc & BLT & SimCSE  & MATCHA  \\ \midrule
50.73 & 35.78 & 50.56 & 50.95 & 34.37 & 9.16 & 59.76 & 31.81  & \textbf{61.08}  \\  \bottomrule
\end{tabular}}}
\caption{Concordance correlation coefficient for metrics on STS-B. MTR, BSc, and BLT denote METEOR, BERTScore, and BLEURT, respectively.}
\label{tab:sts_scores}
\end{table}

\begin{table}[t!]
\setlength\tabcolsep{5pt}
\centering
\scalebox{0.6}{{\renewcommand{\arraystretch}{1.3}
    \centering
    \begin{tabular}{l|ll|ll|ll|ll}
    \toprule
  \multirow{2}{*}{\centering\makecell[c]{\textbf{Metric}}} &  \multicolumn{2}{c|}{\textbf{SNLI}} & \multicolumn{2}{c|}{\textbf{MultiNLI}} & \multicolumn{2}{c|}{\textbf{TruthfulQA}} & \multicolumn{2}{c}{\textbf{Overall}} \\ 
 & \textbf{R@1}  & \textbf{DCG} & \textbf{R@1}  & \textbf{DCG} & \textbf{R@1}  & \textbf{DCG} & \textbf{R@1}  & \textbf{DCG} \\ \midrule
\textbf{R1-F1} & 5.50 & 50.29 & 8.00 & 47.88 & 2.50 & 46.19 & 5.33 & 48.12  \\
\textbf{R2-F1} & 6.00 & 40.94 & 10.50 & 43.53 & 14.00 & 49.00 & 10.17 & 44.49  \\
\textbf{RL-F1} & 1.00 & 48.08 & 3.00 & 45.86 & 3.50 & 47.46 & 2.50 & 47.13  \\ 
\textbf{METEOR} & 11.50 & 47.65 & 11.50 & 46.64 & 9.00 & 48.14 & 10.67 & 47.48  \\ 
\textbf{EmbSim} & 12.50 & 49.05 & 8.00 & 47.00 & 8.00 & 44.06 & 9.50 & 46.71  \\ 
\textbf{BERTScore} & 15.50 & 47.97 & 15.00 & 48.60 & 8.50 & 43.44 & 13.00 & 46.67  \\ 
\textbf{BLEURT} & 6.50 & 39.62 & 4.00 & 40.38 & 13.50 & 48.40 & 8.00 & 42.80  \\ 
\textbf{SimCSE} & 7.00 & 46.74 & 10.50 & 49.13 & 11.00 & 46.35 & 9.50 & 47.40  \\
\textbf{MATCHA} & \textbf{18.50} & \textbf{52.45} & \textbf{23.00} & \textbf{56.08} & \textbf{19.00} & \textbf{50.49} & \textbf{20.17} & \textbf{53.01}  \\\bottomrule
\end{tabular}}}
\caption{Comparison of metric agreement with human judgments. Higher R@1 and DCG are better.}
\label{tab:corr_rank}
\end{table}

\subsubsection{Human Evaluation}\label{subsub:human-eval}

As \citet{liu-etal-2016-evaluate} explains, NLG evaluation metrics must agree with human judgment as their ultimate test. We present a comprehensive human evaluation of MATCHA against the baselines through two complementary studies: (1) a large-scale study on the STS-B dataset, as in previous literature; and (2) a focused study designed to examine how evaluation metrics score contrastive sentence pairs relative to human judgments. Although MATCHA is designed to capture semantic alignment rather than surface similarity, we include STS-B for direct comparability with established baselines. Only for this analysis, we fine-tune MATCHA on the STS-B training split in a separate run that is fully isolated from all other experiments (Table~\ref{tab:sts_scores}).

\textbf{Large-scale human evaluation on STS task.} Prior evaluation metrics like SimCSE~\cite{gao-etal-2021-simcse}, commonly report results on STS, using STS-B as a standard benchmark. Table~\ref{tab:sts_scores} shows that MATCHA attains the highest concordance correlation coefficient with STS-B human annotations, outperforming prior NLG metrics.
These results indicate that MATCHA better captures semantic similarity, with document similarity scores that are most aligned with human judgments.

\textbf{Human evaluation focused on contrastive correct/incorrect pairs.} In order to best demonstrate MATCHA's performance in distinguishing correct from incorrect while also emphasizing false-negative detection, we conduct a focused human study on contrastive pairs. This study is designed to be more focused than the large-scale one, as STS-B does not provide contrastive correct/incorrect pairs.

\textbf{Study Design.}
We randomly sample 200 examples each from test sets of SNLI, MultiNLI, and TruthfulQA, and combine them for Overall evaluation. Human annotators rated the semantic similarity of randomly presented sentence pairs on a 1–5 scale, focusing on semantic meaning rather than surface wording. We recruited eight annotators, primarily scientists from various fields, who were not directly involved in this research. The evaluation interface is publicly available online~\footnote{https://evalmetric.streamlit.app}. Both human ratings and metric scores are linearly scaled to the [0,1] range. To assess alignment with human judgments, we report Rank@1 percentage (R@1), DCG, and concordance correlation coefficient.

\textbf{Ranking Methods by Alignment with Human Judgments.}
As shown in Table~\ref{tab:corr_rank}, MATCHA consistently outperforms all baseline metrics on the overall dataset. It achieves the highest overall Rank@1 score (20.17, +7.17\% over BERTScore), and highest DCG (53.01, +4.89\% over R1-F1). These results suggest that MATCHA produces similarity scores more closely aligned with human perception in ranking quality.

\begin{figure}[t!]
    \centering
    \includegraphics[width=1.0\linewidth]{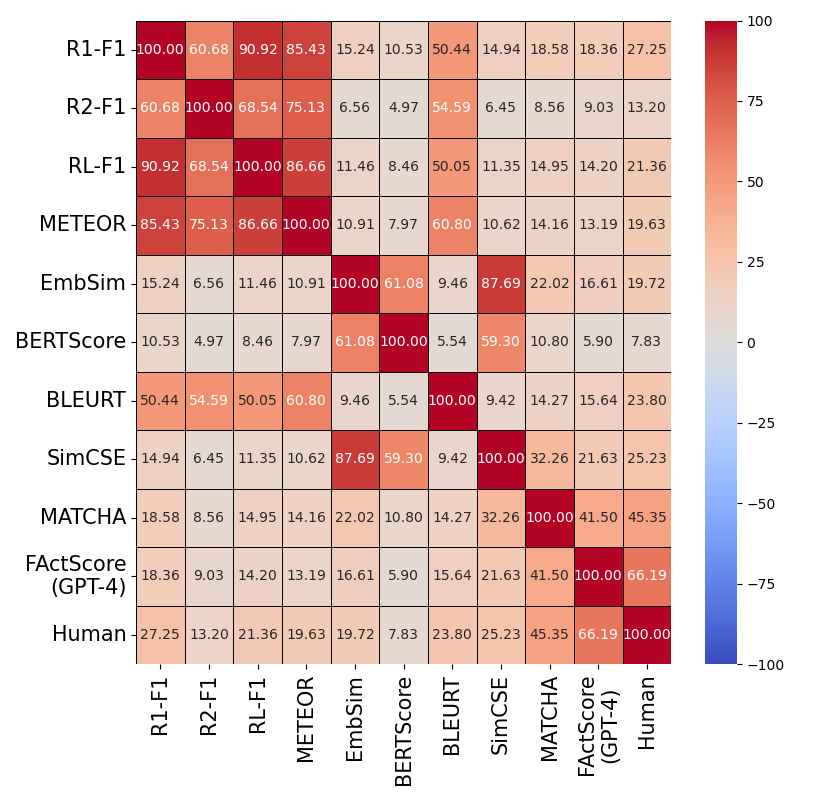}
    \caption{Concordance correlation coefficient among metrics and between metrics and human judgment.}
    \label{fig:ccc_corr}
\end{figure}

\textbf{Correlation with human ratings.}
Figure~\ref{fig:ccc_corr} shows that n-gram metrics (R1-F1, R2-F1, RL-F1, METEOR) correlate strongly with each other (60–90) but poorly with human judgments (13–27), consistent with \citet{liu-etal-2016-evaluate}. Embedding-based methods such as BLEURT (23.80) and SimCSE (25.23) perform better but still struggle with calibration and generalization. Notably, BERTScore yields the weakest correlation with human ratings (7.83), with scores clustered around 0.70-0.80 regardless of quality. In contrast, In contrast, MATCHA achieves the highest correlation with human ratings (45.35), confirming its ability to capture nuanced, human-perceived meaning.

As a further reference, we compare MATCHA to FActScore~\cite{min2023factscore}, a recently proposed metric that evaluates factual accuracy by assessing whether each claim is supported by an external knowledge source. While FActScore achieves a higher correlation (66.19), it is based on an industry-grade LLM (GPT-4~\cite{achiam2023gpt}), with live Internet access, and explicitly optimized for factual consistency rather than semantic similarity, making it an indirect comparison. Despite being much smaller, MATCHA achieves the strongest correlation with FActScore among all metrics, validating its reliability.

\section{Conclusions and Future Work}
\label{sec:conclusion}

Motivated by the limited capability of existing text similarity metrics in distinguishing contradictory or sometimes even semantically unrelated texts from one another, we introduced MATCHA, a contrastive semantic similarity metric which is highly aligned with human judgment. We validate MATCHA via extensive experiments on eight datasets, comparing it with top text similarity metrics across tasks such as question answering, natural language inference, caption generation, summarization, and semantic textual similarity. In conclusion, MATCHA achieves a new state-of-the-art performance with a huge improvement. Additionally, we show that MATCHA can outperform 23 state-of-the-art embedding models in its ability to distinguish correct from incorrect using a reference. The metric is applicable to a wide range of text generation systems, including LLM steering~\cite{braun-etal-2026-beyond}, across dimensions such as topic~\cite{ravenda2025self} and sentiment~\cite{6927566, bahrainian2015sentiment}. Future work may explore extending MATCHA to multilingual and cross-modal contexts, as well as integrating it into training-time objectives to guide generation models toward more semantically faithful outputs.

\section*{Limitations}
\label{app:limitations}

Despite MATCHA’s strong performance and alignment with human judgment across a variety of tasks, it still has some limitations that require further exploration.  

(1) MATCHA has been developed and evaluated primarily on English-language datasets, and its ability to generalize to multilingual contexts remains untested. This may limit its effectiveness in diverse linguistic settings, where variations in grammar, semantics, and cultural nuance can impact similarity judgments.

(2) Although we did not present results using large language models, MATCHA was intentionally designed to remain as small as possible while still achieving strong performance. This decision reflects our consideration that future users of MATCHA may not have access to advanced GPU hardware. While this choice enhances MATCHA’s accessibility and practical utility for the community, it could also be viewed as a limitation, since models with several billion parameters or more were not evaluated.

(3) As with all supervised and learned evaluation metrics, the scoring behavior may reflect biases inherited from the training data and task-specific definitions of correctness. This is a general limitation of data-driven metrics rather than one unique to MATCHA. To reduce this risk, we design MATCHA with broad multi-domain data collection, contrastive supervision across diverse tasks, and an interleaved training strategy that discourages over-specialization to any single benchmark. We further validate its robustness across multiple benchmarks and domains, although no trained metric can fully account for all possible linguistic variations and contextual notions of correctness.

\section*{Acknowledgments}
The authors thank the International Max Planck Research School for Intelligent Systems (IMPRS-IS) for their support.

\bibliography{custom}

\appendix

\section{Experimental Details}
\label{sec:appendix}

\subsection{Training Details.}\label{app:training-details}

For MATCHA, we initialize the embedding layer with pre-trained word embedding layer from GPT-2 (GPT2-small). For supervised training, we use the training sets from SNLI~\cite{bowman-etal-2015-large}, MultiNLI~\cite{williams-etal-2018-broad}, Climate-Fever~\cite{diggelmann2020climatefever}, NEWTS~\cite{bahrainian-etal-2022-newts} VitaminC~\cite{schuster-etal-2021-get}, AltLex~\cite{hidey2016identifying}, SimpleWiki~\cite{coster2011simple}, Sentence Compression~\cite{filippova2013overcoming}, WikiHow~\cite{koupaee2018wikihow}, STS-B (English)~\cite{cer-etal-2017-semeval} and 100k randomly sampled examples each (to ensure more balanced dataset sizes) from COCO-Caption~\cite{lin2015microsoft} training dataset, Flickr30k~\cite{plummer2015flickr30k}, QQP~\cite{quora-question-pairs}, ParaNMT~\cite{wieting-gimpel-2018-paranmt}, and WikiAnswers~\cite{Fader14}. We apply interleaved training~\cite{kirkpatrick2017overcoming}, where each batch is sampled from a different dataset, to mitigate catastrophic forgetting across diverse tasks. For the contrastive learning process, we use contrastive pairs from SNLI, MNLI, VitaminC, and QQP; for the other datasets, we construct incorrect examples by sampling a random item as the incorrect candidate. In total, we train on approximately 1 million (reference, correct candidate, incorrect candidate) triplets across 15 integrated datasets. 

\subsection{Evaluation Datasets}\label{app:eval-datasets}
SNLI and MultiNLI serve as NLI benchmarks, using entailments as correct pairs and contradictions as incorrect pairs, with \textasciitilde5.9k validation pairs each. TruthfulQA is a question-answering dataset consisting of questions intentionally crafted to elicit common misconceptions. It provides 1k validation pairs from 500 questions, where each pair consists of the correct and incorrect answers, which we use as our samples. COCO-Caption is derived from the MS-COCO dataset, a benchmark for evaluating image captioning systems. It includes 1k randomly selected samples; for each, four concatenated captions form the reference, one additional caption serves as the correct output, and an irrelevant caption is used as the incorrect output, yielding 2k validation pairs. In the NEWTS topic-based summarization dataset, we use the full summary as the reference, the topic sentence as the correct instance, and an unrelated sentence as the incorrect one, yielding \textasciitilde2.4k validation pairs. Additionally, STS-B is a large-scale Semantic Textual Similarity dataset and comprises \textasciitilde1.4k test sentence pairs annotated with human similarity scores.

\subsection{Implementation Details.}\label{app:implementation}

We train MATCHA on 8 NVIDIA A100-PCIE-40GB GPUs with a batch size of 128 and gradient accumulation performed every 8 steps. The model is trained for 15 epochs using the Adam optimizer with a learning rate of 1e-4. We use a weight decay of 0.05 and apply exponential learning rate decay after each epoch. We set the random seed to 42 for reproducibility. 
The input sequence length is 512, and the context vector dimension $N_c$ is 16.

\section{Semantic Separation in LM Embedding Spaces}\label{app:lm-seperation}

\begin{figure*}[t!]
    \centering
    \includegraphics[width=\textwidth]{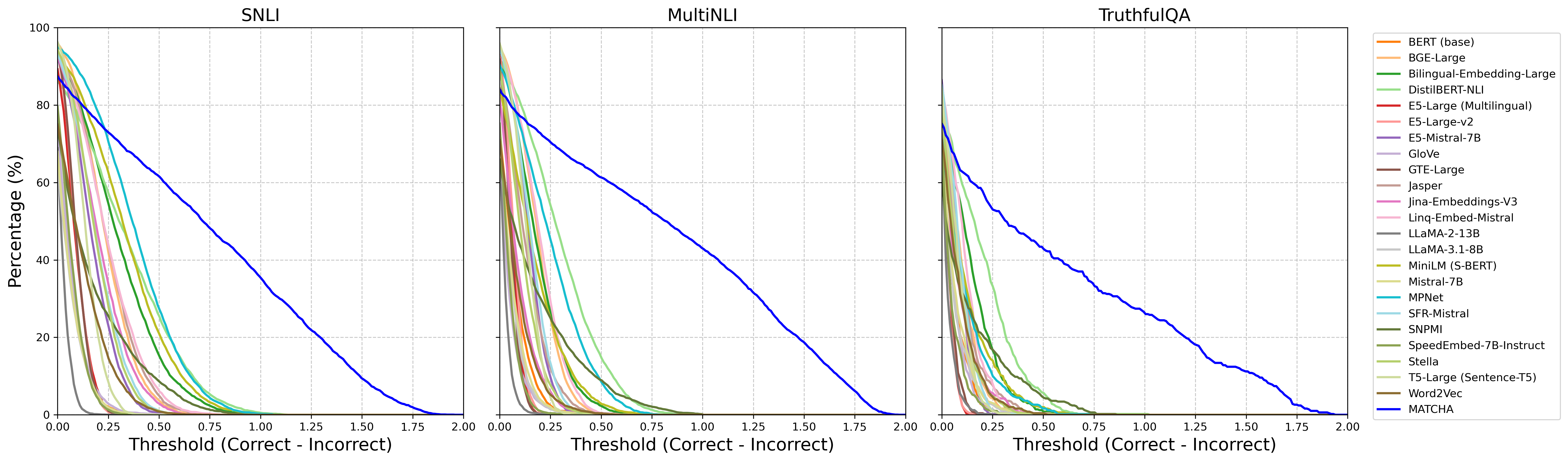}
    \caption{Comparison of embedding similarity gaps across embedding models on SNLI, MultiNLI, and TruthfulQA. The y-axis shows the percentage of samples satisfying $(\text{Sim}_C - \text{Sim}_I) > \text{Threshold}$.}\label{fig:embedding_gap}
\end{figure*}

\textbf{Semantic separation across thresholds.} The threshold curves in Figure \ref{fig:embedding_gap} further illustrate MATCHA’s advantage in maintaining semantic separation, where document representations are constructed by transformer and LLM-based models using mean pooling over final hidden states, while static embeddings (e.g., GloVe, Word2Vec) use mean pooling over token vectors. A zero threshold corresponds to cases where correct and incorrect candidates receive nearly identical similarity scores, often reflecting model uncertainty. Scores within a threshold of 0.1 can be viewed as a “jitter region,” where differences are too small to be semantically meaningful. Many embeddings (e.g., MPNet, T5-Large) perform well at these very low distinction thresholds but collapse rapidly as the threshold increases, especially on the unseen-domain TruthfulQA dataset. Notably, Jina-Embeddings-V3 and SFR-Embedding-Mistral, while being top 10 in STS from the MTEB benchmark, retain modest separation at low thresholds but rapidly lose semantic differentiation. Even DistilBERT-NLI, which was fine-tuned on NLI datasets, degrades rapidly on TruthfulQA. Instead, MATCHA maintains a large and stable gap between correct and incorrect pairs in all datasets, with performance that decreases smoothly and consistently as the threshold rises.

\begin{figure*}[t!]
    \centering

    \begin{subfigure}[b]{1\textwidth}
        \includegraphics[width=\textwidth]{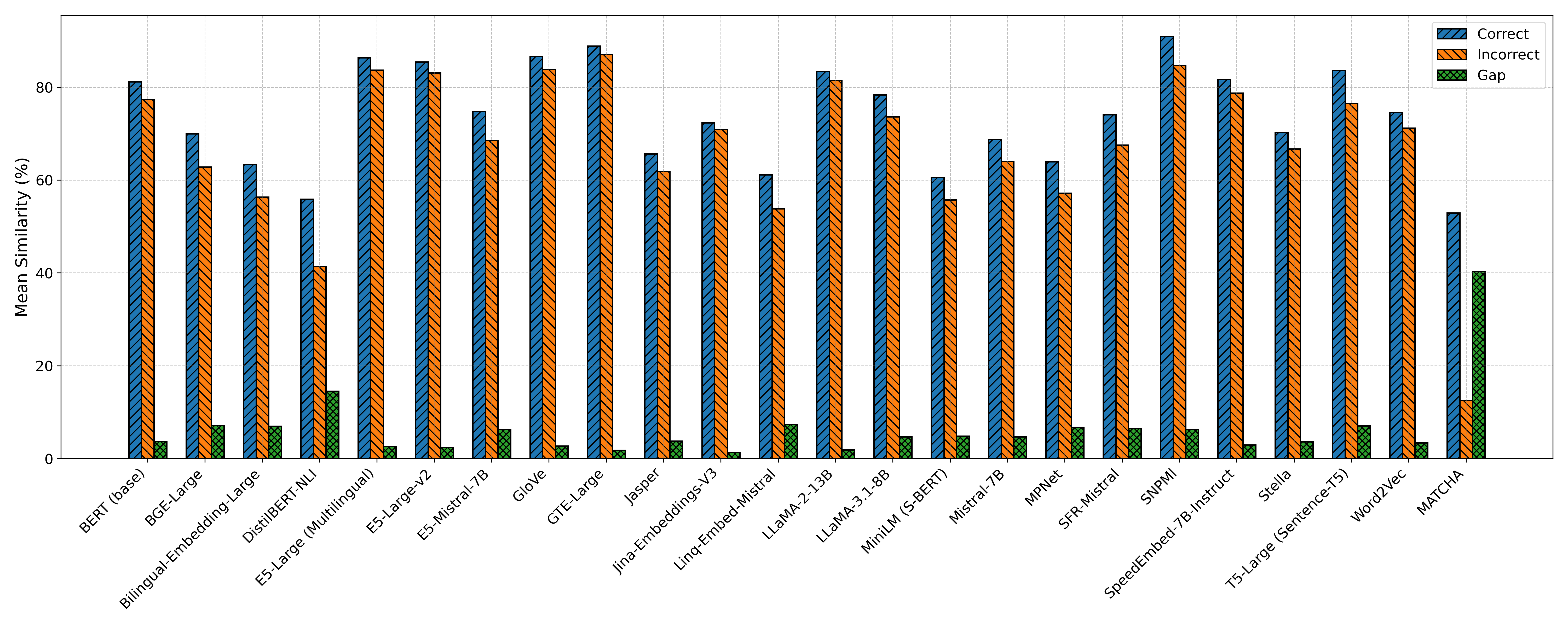}
        \caption{Climate-Fever}
        \label{fig:avg_lmgap_climate}
    \end{subfigure}
    \vspace{0.5cm} 
    
    \begin{subfigure}[b]{1\textwidth}
        \includegraphics[width=\textwidth]{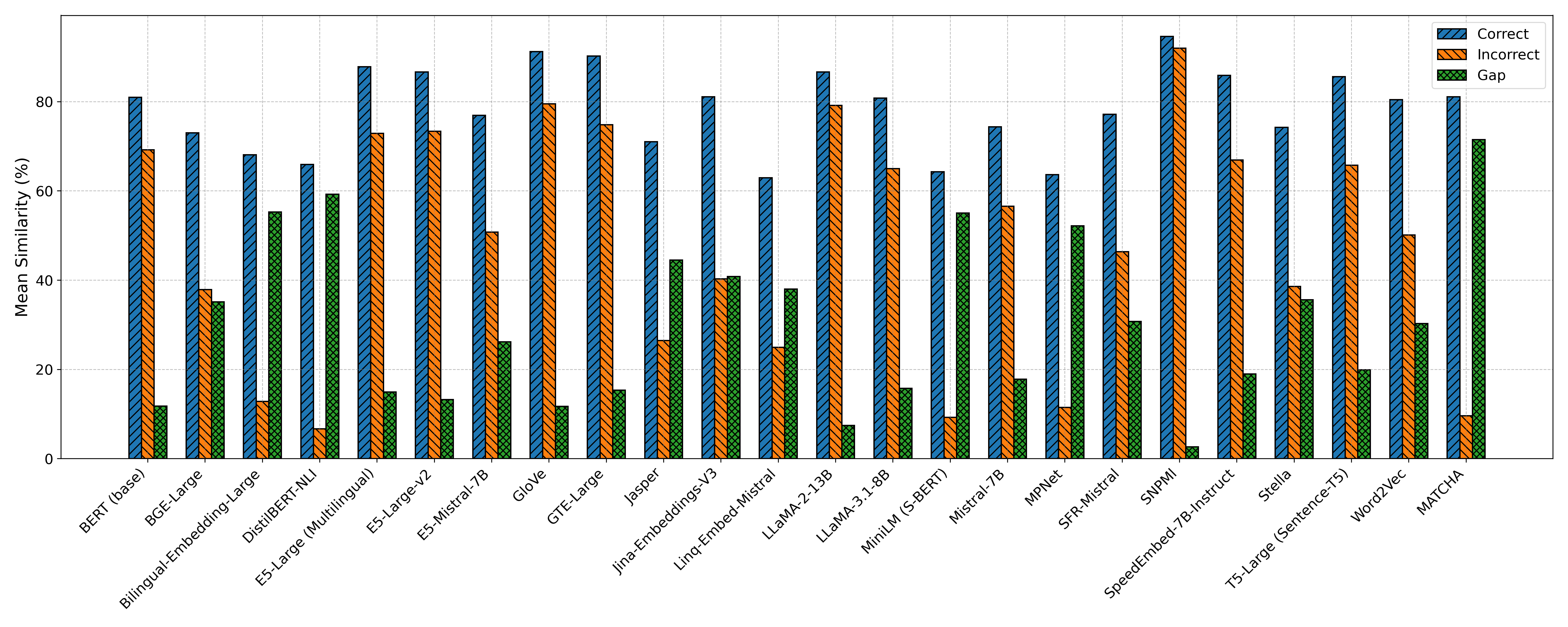}
        \caption{COCO-Caption}
        \label{fig:avg_lmgap_coco}
    \end{subfigure}

    \vspace{0.5cm} 

    \begin{subfigure}[b]{1\textwidth}
        \includegraphics[width=\textwidth]{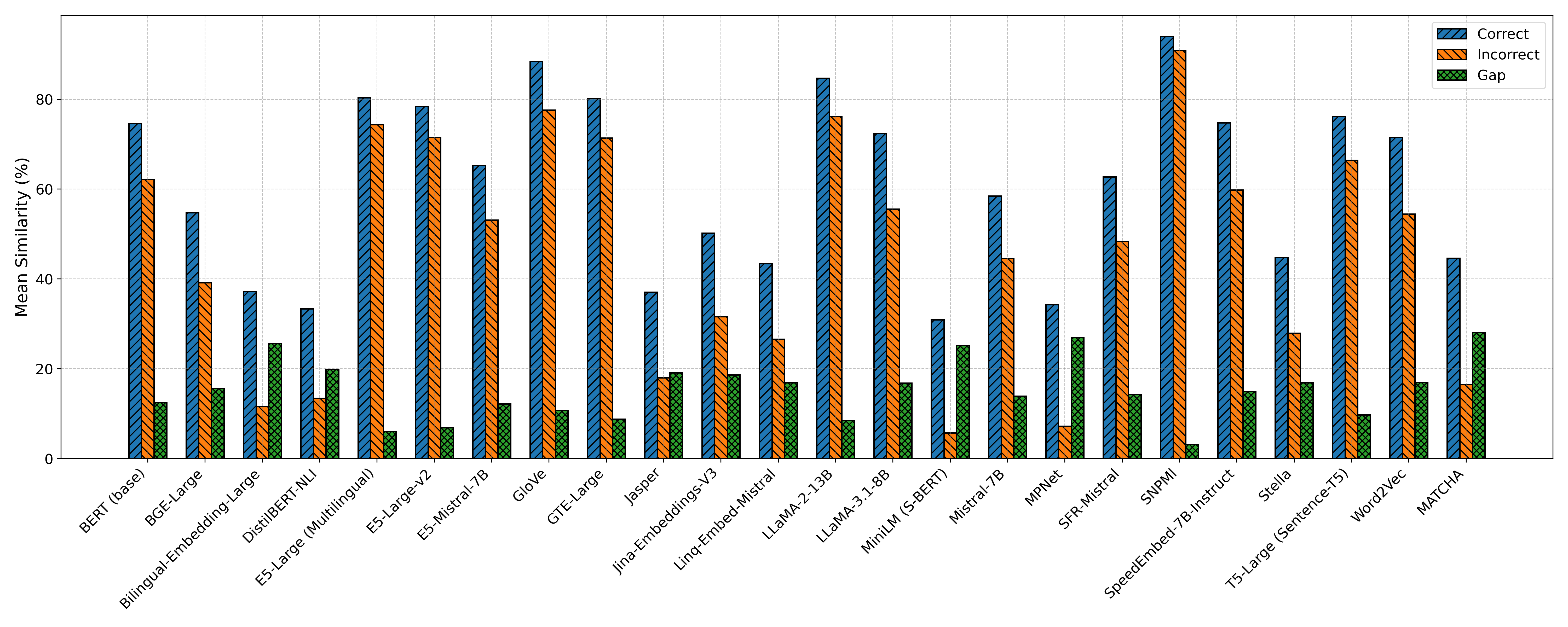}
        \caption{NEWTS}
        \label{fig:avg_lmgap_newts}
    \end{subfigure}

\caption{Average similarity scores and gaps between correct and incorrect pairs across embedding models on Climate-Fever, COCO-Caption, and NEWTS.}
\label{fig:app_avg_lmgap}
\end{figure*}

\begin{figure*}[t!]
    \centering

    \begin{subfigure}[b]{0.47\textwidth}
        \includegraphics[width=\textwidth]{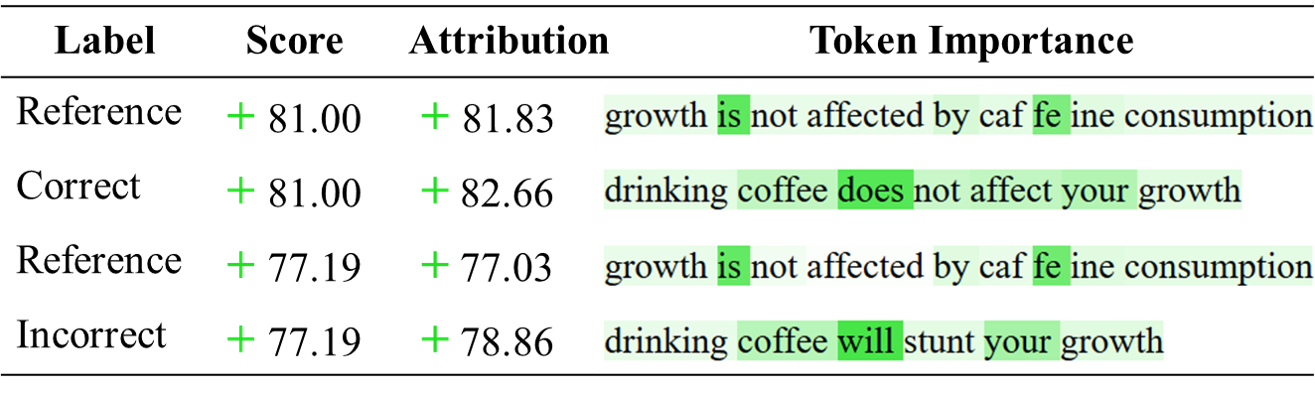}
        \caption{EmbSim}
        \label{fig:transformer_token}
    \end{subfigure}
    \hspace{0.02\textwidth}
    \begin{subfigure}[b]{0.47\textwidth}
        \includegraphics[width=\textwidth]{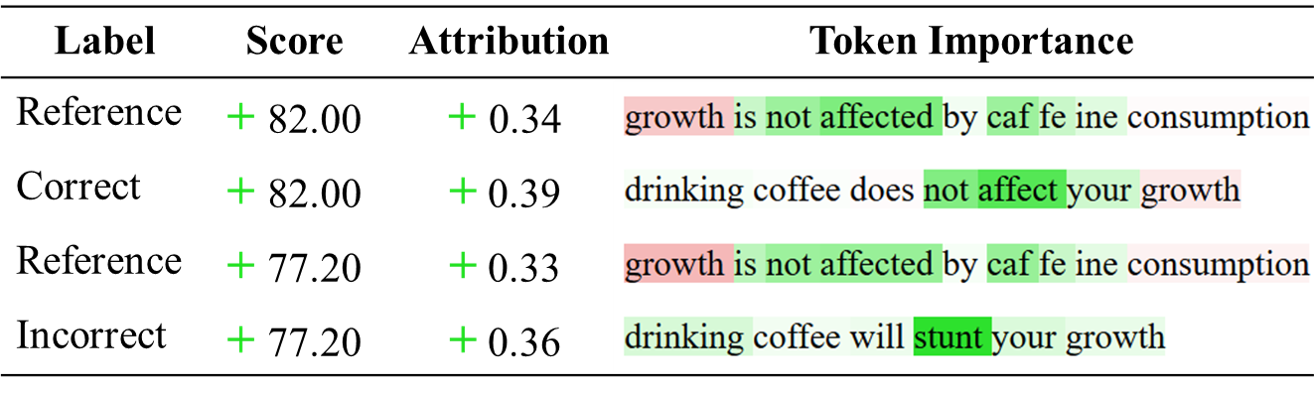}
        \caption{BERTScore}
        \label{fig:bertscore_token}
    \end{subfigure}

    \vspace{0.3cm} 

    \begin{subfigure}[b]{0.47\textwidth}
        \includegraphics[width=\textwidth]{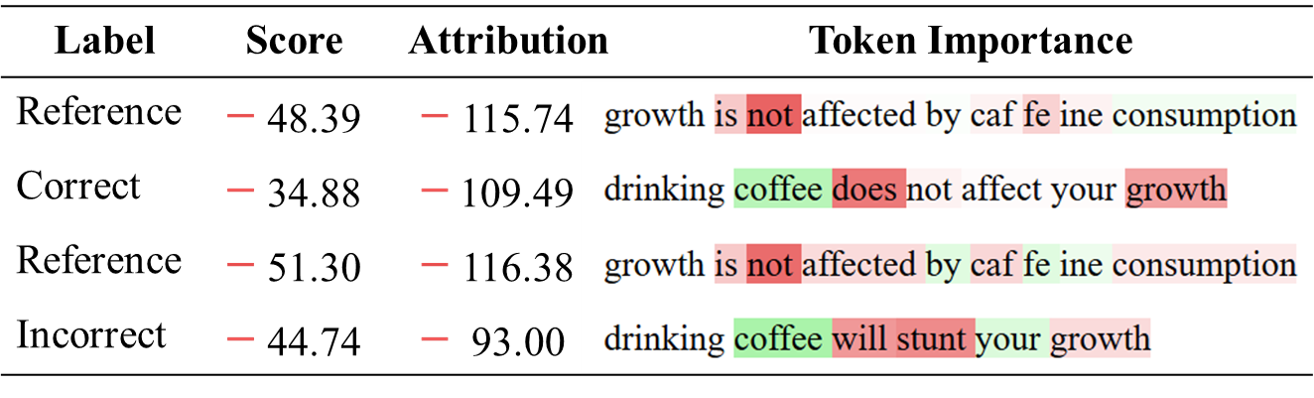}
        \caption{BLEURT}
        \label{fig:bleurt_token}
    \end{subfigure}
    \hspace{0.02\textwidth}
    \begin{subfigure}[b]{0.47\textwidth}
        \includegraphics[width=\textwidth]{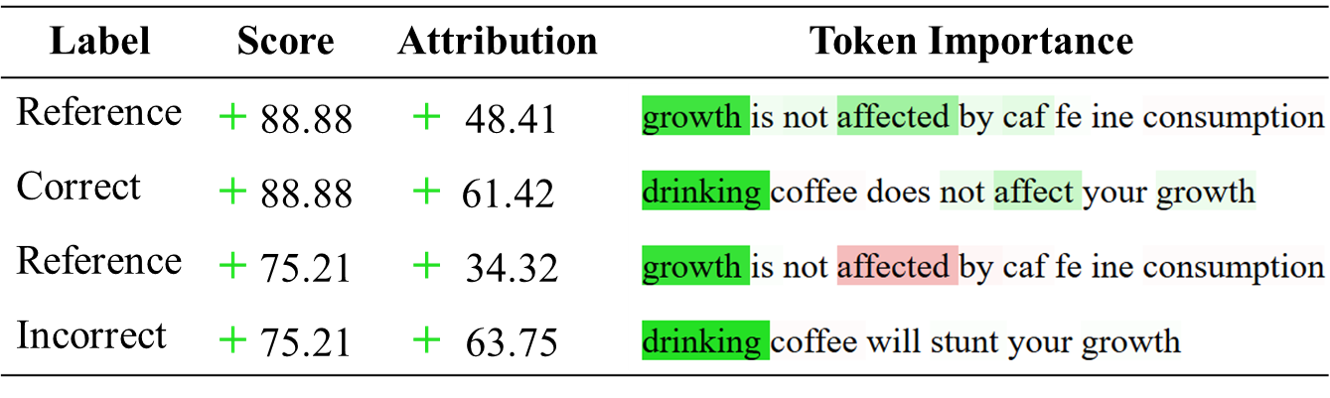}
        \caption{SimCSE}
        \label{fig:simcse_token}
    \end{subfigure}
    
    \vspace{0.3cm}

    \begin{subfigure}[b]{0.47\textwidth}
        \includegraphics[width=\textwidth]{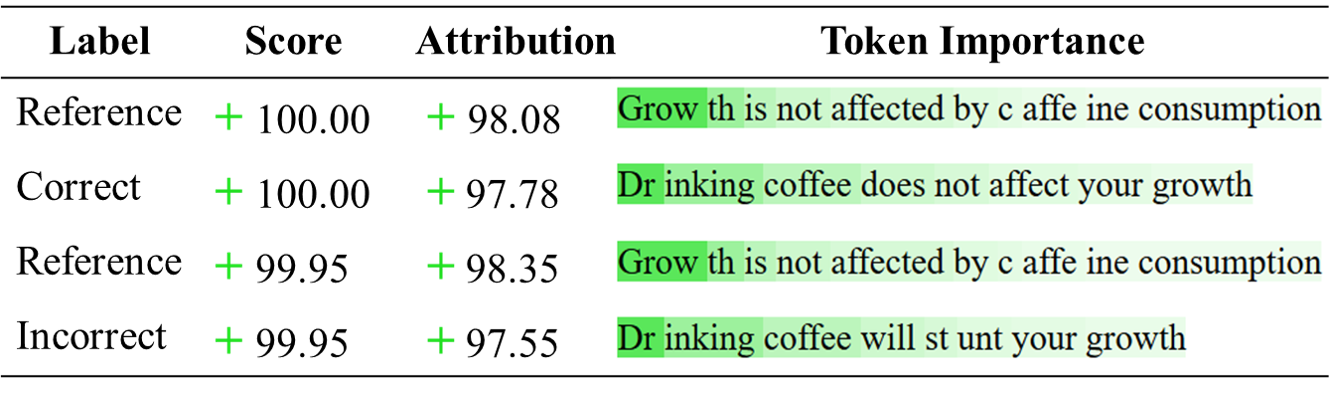}
        \caption{Mistral-7B}
        \label{fig:mistral_token}
    \end{subfigure}
    \hspace{0.02\textwidth}
    \begin{subfigure}[b]{0.47\textwidth}
        \includegraphics[width=\textwidth]{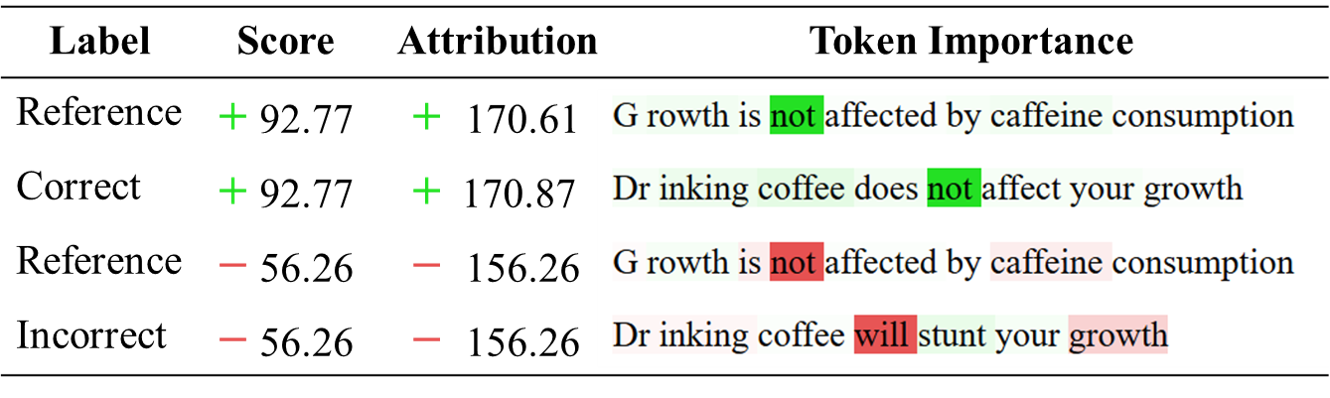}
        \caption{MATCHA}
        \label{fig:MATCHA_token}
    \end{subfigure}
    \label{fig:token_weights}
\caption{Token-level importance visualization for pairs of (Reference, Correct) and (Reference, Incorrect) sentences. \textcolor{green}{Green} highlights indicate tokens contributing positively to the similarity score, while \textcolor{red}{red} highlights indicate conflicting tokens contributing negatively. Scores are from each metric, and the attribution scores reflect the total contribution of all tokens to the final metric score.}
\label{fig:token_importance}
\end{figure*}

\section{Token-Level Interpretability}\label{app:token-interpret}

\begin{table}[t!]
\setlength\tabcolsep{5pt}
\centering
\scalebox{0.6}{{\renewcommand{\arraystretch}{1.3}
\begin{tabular}{l|ll|l|ll}
\toprule
\multirow{2}{*}{\centering\makecell[c]{\textbf{Model}}} &  \multicolumn{2}{c|}{\textbf{Attribution}} & \multirow{2}{*}{\centering\makecell[c]{\textbf{Model}}} &  \multicolumn{2}{c}{\textbf{Attribution}}  \\ 
 & \textbf{(Corr, Incorr)} & \textbf{Gap}  & & \textbf{(Corr, Incorr)} & \textbf{Gap} \\ \midrule
\textbf{EmbSim} & (66.80, 47.82) & 18.98 & \textbf{BERTScore} & (0.42, 0.40) & 0.02 \\ \textbf{BLEURT} & (-1.78, -76.37) & 74.59 & 
 \textbf{SimCSE} & (79.84, 83.09) & -3.25 \\ \textbf{Mistral-7B} & (95.88, 95.85) & 0.03  & \textbf{MATCHA} & \textbf{(60.63, -41.80)} & \textbf{102.43} \\ \bottomrule
\end{tabular}}}
\caption{Average attribution scores for correct and incorrect pairs (600 samples), along with their gap.}
\label{tab:attribution}
\end{table}

\begin{table*}[t!]
\setlength\tabcolsep{4pt}
\centering
\scalebox{0.85}{{\renewcommand{\arraystretch}{1.3}
\begin{tabular}{l|ccccccc}
\toprule
\textbf{Training Setup} & \textbf{SNLI} & \textbf{MNLI} & \textbf{TruthfulQA} & \textbf{Climate} & \textbf{COCO} & \textbf{NEWTS} & \textbf{CCC} \\
\midrule
Contrastive Triplets (5D) & 34.13 & 34.72 & \textbf{23.37} & 17.33 & 23.30 & 9.10 & 42.69 \\
Random Negative (10D) & 13.83 & 3.52 & 3.91 & 4.19 & 42.46 & \textbf{21.70} & 13.70 \\
Sequential Training (15D) & 26.32 & 27.70 & 16.79 & 10.42 & \textbf{45.24} & 20.30 & 36.01 \\
Curriculum Learning (15D) & 13.36 & 28.61 & 22.48 & 16.71 & 27.91 & 14.19 & 34.27 \\
\textbf{Interleaved Training (15D)} & \textbf{34.95} & \textbf{37.27} & \textbf{23.33} & \textbf{20.17} & 35.76 & 14.05 & \textbf{45.35} \\
\bottomrule
\end{tabular}}}
\caption{Ablation study on training data composition, negative sampling, and optimization strategy. We report the Wasserstein distance between correct and incorrect score distributions across six evaluation benchmarks, together with the overall concordance correlation coefficient (CCC). D denotes the number of datasets used during training. Interleaved training consistently yields the strongest and most balanced performance across domains.}
\label{tab:ablation_training}
\end{table*}

As a qualitative analysis, we compute token-level attributions with Integrated Gradients on 600 pairs randomly sampled from SNLI, MultiNLI, and TruthfulQA (200 each). Positive attributions increase the similarity score (\textcolor{green}{green}); negative attributions decrease it (\textcolor{red}{red}). Figure~\ref{fig:token_importance} visualizes two sentence pairs, (Reference, Correct) and (Reference, Incorrect), with bidirectional analysis shown as four rows per metric.

EmbSim, BERTScore, BLEURT, and SimCSE emphasize general content words (e.g., \textit{“caffeine”}, \textit{“growth”}) and fail to flag contradiction-relevant tokens, producing nearly identical scores for correct and incorrect pairs.

Mistral-7B embedding model from Section~\ref{subsub:embedding_comparison} gives uniformly high similarity scores ($>$0.9) regardless of alignment. As shown in Figure~\ref{fig:mistral_token}, attribution weights decay from left to right, reflecting the next-token prediction bias of autoregressive LLMs rather than semantic distinctions.

By contrast, MATCHA captures nuanced semantic differences between aligned and contradictory pairs. It assigns strong positive attributions to meaning-preserving tokens such as \textit{“not”} and \textit{“does not”}, while giving strong negative attributions to contradiction cues like \textit{“will”}. This fine-grained token sensitivity enables MATCHA, unlike other metrics, to successfully distinguish contradictory inputs. To assess whether this behavior holds at scale, we now move from these examples to dataset-wide results.

For a broader view, Table~\ref{tab:attribution} summarizes average token-level attributions for correct versus incorrect pairs. BERTScore, SimCSE, and Mistral-7B embedding model yield nearly identical averages for the two conditions, indicating limited sensitivity to contradiction. EmbSim shows a modest gap (18.98). BLEURT reports a larger numerical gap (74.59), but both means are negative, making its attributions harder to interpret. MATCHA achieves the largest attribution gap (102.43) between correct and incorrect pairs, highlighting its ability to isolate meaning-bearing differences for humans and provide token-level interpretability.

\section{Ablation on Training Data Composition and Optimization Strategy}\label{app:ablation-study}

To better understand which training choices contribute most to MATCHA’s final performance, we conduct an ablation study over dataset composition, negative sampling strategy, and optimization schedule. Table~\ref{tab:ablation_training} reports the Wasserstein distance between correct and incorrect score distributions across six evaluation benchmarks, together with the overall concordance correlation coefficient (CCC).

The Contrastive Triplets (5D) setting, trained on approximately 300k contrastive triplets, performs strongly on entailment-sensitive and factuality-oriented datasets such as SNLI, MNLI, and
TruthfulQA, but its narrower dataset diversity limits broader generalization.

The Random Negative (10D) setting, trained on approximately 700k automatically constructed negatives, performs better on surface-level generation benchmarks such as COCO-Caption and NEWTS, yet struggles to reliably distinguish semantically similar contradictions.

Scaling to 15 datasets with Sequential Training improves overall robustness, highlighting the importance of multi-domain coverage. Curriculum Learning (15D) yields moderate gains but remains sensitive to dataset ordering effects.

The strongest and most consistent performance is achieved with Interleaved Training (15D), which combines broad dataset diversity with structured contrastive supervision while preventing dominance by any single domain. This configuration achieves the highest overall CCC (45.35) and the strongest Wasserstein distances on most benchmarks, thereby motivating our final choice of interleaved multi-domain training.

Overall, these findings suggest that MATCHA’s gains arise primarily from balanced multi-domain contrastive learning, rather than simply scaling training data volume.

\end{document}